\documentclass[letterpaper]{article} 
\usepackage{aaai2026}  
\usepackage{times}  
\usepackage{helvet}  
\usepackage{courier}  
\usepackage[hyphens]{url}  
\usepackage{graphicx} 
\usepackage{natbib}  
\usepackage{caption} 
\usepackage{booktabs}
\usepackage[utf8]{inputenc} 
\usepackage[T1]{fontenc}    
\usepackage{amsfonts}       
\usepackage{nicefrac}       
\usepackage{microtype}      
\usepackage{xcolor}         
\usepackage{amsmath} 
\usepackage{float}  
\usepackage{mathtools}
\usepackage{subcaption}

\title{
Faster by Design: Interactive Aerodynamics via Neural Surrogates Trained on Expert-Validated CFD
}
\author{
    Nicholas Thumiger\textsuperscript{\rm 1},
    Andrea Bartezzaghi\textsuperscript{\rm 1},
    Mattia Rigotti\textsuperscript{\rm 1},
    Cezary Skura\textsuperscript{\rm 1},
    Thomas Frick\textsuperscript{\rm 1},\\
    Elisa Serioli\textsuperscript{\rm 2},
    Fabrizio Arbucci\textsuperscript{\rm 2},
    A. Cristiano I. Malossi\textsuperscript{\rm 1}
}
\affiliations{
    \textsuperscript{\rm 1}IBM Research\\
    \textsuperscript{\rm 2}Dallara Automobili
}

\begin{document}

\maketitle

\begin{abstract}
Computational Fluid Dynamics (CFD) is central to race-car aerodynamic development, yet its cost --- tens of thousands of core-hours per high-fidelity evaluation --- severely limits the design space exploration feasible within realistic budgets.
AI-based surrogate models promise to alleviate this bottleneck, but progress has been constrained by the limited complexity of public datasets, which are dominated by smoothed passenger-car shapes that fail to exercise surrogates on the thin, complex, highly loaded components governing motorsport performance.
This work presents three primary contributions.
First, we introduce a high-fidelity RANS dataset built on a parametric LMP2-class CAD model and spanning six operating conditions (map points) covering straight-line and cornering regimes, generated and validated by aerodynamics experts at Dallara to preserve features relevant to industrial motorsport.
Second, we present the Gauge-Invariant Spectral Transformer (GIST), a graph-based neural operator whose spectral embeddings encode mesh connectivity to enhance predictions on tightly packed, complex geometries.
GIST guarantees discretization invariance and scales linearly with mesh size, achieving state-of-the-art accuracy on both public benchmarks and the proposed race-car dataset.
Third, we demonstrate that GIST achieves a level of predictive accuracy suitable for early-stage aerodynamic design, providing a first validation of the concept of interactive design-space exploration --- where engineers query a surrogate in place of the CFD solver --- within industrial motorsport workflows.
\end{abstract}

\section{Introduction}


Computational Fluid Dynamics (CFD) sits at the core of modern race-car aerodynamic development, driving the iterative loop through which teams extract aero metrics such as drag and downforce under tight regulatory and budget constraints.
Yet, CFD is computationally expensive: a single high-fidelity steady-state evaluation of a full car typically consumes thousands of core-hours, and meaningful design exploration requires sweeping both geometric parameters --- such as wing profiles, endplates, underfloor geometry --- and operating conditions, i.e., the map points defining heave, pitch, yaw, roll and steer.
The combinatorial cost of this exploration is a fundamental bottleneck of motorsport aerodynamic design, directly limiting the number of concepts feasible to evaluate within time and resource budget.

AI-based surrogate models offer the opportunity to accelerate design cycles by replacing the numerical solver with a learned approximator that predicts flow fields and integrated loads at a fraction of the cost of a CFD run  \citep{azizzadenesheli2024neural}.
Within this space, data-driven approaches, which learn directly from precomputed simulation data rather than embedding the governing equations into the training objective, have emerged as the most practical route for the high-dimensional geometrically complex configurations encountered in industrial aerodynamics.
Neural operators in particular have substantially raised the ceiling of surrogate accuracy on PDE-governed problems \citep{kovachki2023neural}.

However, in the automotive domain, progress has been constrained by the limited complexity of publicly available CFD datasets. Existing benchmarks are dominated by smoothed passenger-car shapes or academic bluff bodies, none of which contain the thin, complex, highly loaded components (e.g.~wings, endplates, strakes, splitters, diffusers) that dominate race-car aerodynamics. Surrogates trained or validated on such data are therefore not exercised in the regimes that matter for motorsport. Moreover, numerical solutions available in the public domain are often not validated by aerodynamics experts.
Additionally, existing datasets are typically generated at a single fixed operating condition, whereas industrial aerodynamic development requires sweeping map points --- combinations of heave, pitch, yaw, roll, and steer --- to characterize vehicle behaviour across the full range of track conditions.

This work presents the following complementary contributions: (i)~a high-fidelity CFD dataset spanning six map points across straight-line and cornering conditions, built with the purpose to expose surrogate models to race-car-representative geometry and flow features, and developed and validated by aerodynamics domain experts at Dallara, (ii)~GIST, a graph-based neural operator model designed to operate on the graphs that arise from such complex geometries, and to scale to the mesh sizes such problems entail, and (iii)~empirical evidence that GIST's accuracy on the race-car dataset is sufficient for early-stage aerodynamic design, pointing toward its adoption within industrial workflows.

We firstly present the dataset, then we briefly introduce the method, and finally we present results obtained on the race-car dataset and an example of how the surrogate model can aid optimization of geometrical design.

\section{Methodology}

\subsection{Dataset}\label{sec:dataset}



The dataset is generated from a simplified parametric CAD model of a Le Mans Prototype 2 (LMP2) race car, designed in-house by aerodynamics domain experts at Dallara to retain the features that govern vehicle-scale aerodynamics: for example, the rotating wheels encapsulated into the wheelhouse create complex wakes, which interact with the complete underfloor, from front splitter to rear diffuser, in ground effect (Venturi behaviour). To our knowledge, no public CFD dataset on cars combines this level of aerodynamic detail with expert-driven generation and validation.

Each geometric configuration of the parametric CAD model is evaluated at six map points, reproducing both straight-line and cornering track conditions through different combinations of heave, pitch, yaw, roll, and steer.
To our knowledge, no existing public automotive CFD dataset covers multiple operating conditions in this way: standard benchmarks fix a single flow condition, which precludes training surrogates capable of predicting aerodynamic behaviour across the map-point space central to race-car development.
The car surface is decomposed into 20 part identifiers (PIDs), each corresponding to a distinct aerodynamic component or to special surfaces (to be treated accordingly). Individual PIDs can be treated separately, enabling component-level analysis, or merged into a single watertight surface mesh for full-car surrogate training and inference.
The geometric parameters are chosen specifically because they induce large, non-local changes in the flow: e.g., a modification on the top hole gurney of the front wheelhouse reshapes the wheel wake, feeding  the underfloor differently, thus producing strong load variations on components that were not themselves modified. The dataset is therefore designed to stress a surrogate model's ability to capture both the direct effect (loads and surface flow features on the parametrized component) and the indirect effect (load redistribution across the remainder of the car), as typical of the elliptic field defined by the incompressible Navier-Stokes equations. For race-car aerodynamics, where global aerodynamic balance is at least as important as local peak loads, this distinction is essential.

The geometry also includes a radiator modeled as a porous medium with a single inlet and outlet (marked as different PIDs), where a pressure drop across the flow direction is imposed following Darcy's law. This captures the effect of the cooling system on the external aerodynamics, a coupling that is routinely neglected in public datasets yet materially affects global downforce and drag.

CFD simulations are performed as incompressible steady-state RANS \citep{pope2000turbulent}:

\begin{equation*}
\frac{\partial \bar{u}_i}{\partial x_i} = 0
\end{equation*}

\begin{equation*}
\overline{u}_j \frac{\partial \overline{u}_i}{\partial x_j} = -\frac{1}{\rho} \frac{\partial \overline{p}}{\partial x_i} + \nu \frac{\partial^2 \overline{u}_i}{\partial x_j \partial x_j} - \frac{\partial}{\partial x_j} (\overline{u'_i u'_j}),
\end{equation*}
where the Reynolds stress tensor $(\overline{u'_i u'_j})$ is modeled according to the Boussinesq hypothesis, which relates the stress to the mean strain rate:
\begin{equation*}
-\overline{u'_i u'_j} = \nu_t \left( \frac{\partial \overline{u}_i}{\partial x_j} + \frac{\partial \overline{u}_j}{\partial x_i} \right) - \frac{2}{3} k \delta_{ij}.
\end{equation*}
The $k-\omega$ SST turbulence model \citep{menter1992improved}, properly customized by Dallara experts for correlation purposes, is chosen to model the kinematic turbulent viscosity $\nu_t$.

The mesh has been created according to the best practices of Dallara experts, preserving geometric features with high-quality elements, realizing a fine modeling of the boundary layer and of the volumetric regions where strong gradients of the solution happen.
Second order spatial discretization on the velocity-pressure coupled solver guarantees high accuracy, with convergence evaluated on both residuals and aerodynamics metrics.
Target fields stored on the car surface for each sample include mean pressure and mean wall shear stress along the three directional components. Together, these provide sufficient information for both load integration and surface-flow diagnostics.

Every configuration has been reviewed and aero-verified by Dallara aerodynamic engineers against expected load trends, providing an expert-in-the-loop validation step grounded in motorsport domain expertise.
A summary of the key dataset parameters is provided in Table~\ref{tab:dataset}.
Extensions to the dataset are currently in progress, including additional map points and a substantial refinement of the geometry.

\begin{table}[!tbp]
\centering
\caption{LMP2 race car dataset sample characteristics.}
\label{tab:dataset}
\begin{tabular}{lr}
\toprule
Average points per sample & $\sim 4.8\text{M}$ \\
Average cells per sample & $\sim 2.4\text{M}$ \\
Average sample size & $360 \text{ MB}$ \\
\midrule
Map points per configuration & 6 \\
Operating conditions & heave, pitch, yaw, roll, steer \\
\midrule
Total samples & 625 \\
\bottomrule
\end{tabular}
\end{table}

\subsection{Neural operators}

\paragraph{Evolution of Neural Operators}
Two foundational frameworks established the operator-learning paradigm: FNO \citep{li2020fourier}, which applies spectral convolutions in Fourier space, and DeepONet \citep{lu2021learning}, which approximates operators via branch-trunk network pairs.
Early data-driven CFD methods built on these ideas: graph network simulators such as MeshGraphNet \citep{pfaff2021learning} demonstrated that learned message passing on unstructured meshes could reproduce complex fluid dynamics, while the Geometry-Informed Neural Operator (GINO, \citealt{li2023geometry}) combined graph encoders with FNO-based architectures to handle 3D car shapes.
As the field transitioned toward transformer-based architectures, standard attention mechanisms presented a severe quadratic scalability problem when applied to the massive, high-resolution meshes required for vehicle aerodynamics.
Transolver mitigates this scalability bottleneck by projecting spatial coordinates onto a reduced set of context tokens.
However, this efficiency inherently sacrifices the faithful encoding of the high-fidelity mesh structure, limiting the model's ability to generalize reliably across varying grid resolutions \citep{wu2024transolver}.
The Geometry Aware Operator Transformer (GAOT, \citealt{wen2025gaot}) similarly combines an attentional graph neural operator encoder with a vision transformer processor and geometry-aware embeddings, achieving strong performance across diverse PDE benchmarks.
Like Transolver, however, GAOT relies on Euclidean proximity for graph construction, making it susceptible to unphysical cross-surface interactions on thin-walled geometries.

\paragraph{GIST}
To resolve the scalability limitations of existing architectures without compromising geometric fidelity, we introduce the Gauge-Invariant Spectral Transformer (GIST), detailed in \citet{rigotti2026gist}.

A surface mesh discretizing the car's CAD geometry is a graph $G = (V, \mathcal{E})$, where vertices $V$ are the mesh nodes and edges $\mathcal{E}$ encode their connectivity.
Treating meshes as graphs rather than unstructured point clouds is critical: it preserves topological information that Euclidean proximity alone cannot capture, such as the fact that two vertices on opposite sides of a thin wing are geometrically close but topologically distant.
A neural operator is discretization-invariant if its predictions are consistent across different mesh resolutions and samplings of the same physical surface, a property that most architectures violate in practice because their node features shift non-trivially as the mesh changes.
Spectral methods offer a principled route through the normalized graph Laplacian $L = I - D^{-1}A$ (where $A$ is the adjacency matrix of $G$ and $D$ is the diagonal degree matrix), whose spectrum converges to that of the continuous Laplace-Beltrami operator as mesh resolution increases.
However, spectral methods face two compounding obstacles in this setting.
First, computing the eigenvectors of $L$ via full eigendecomposition costs $\mathcal{O}(N^3)$ for a mesh with $N$ nodes, which is entirely intractable at the ${\sim}4.8$M-node scale of our dataset.
Second, eigenvectors of $L$ suffer from gauge ambiguity: any eigenvector $u_k$ may be arbitrarily sign-flipped or rotated within degenerate eigenspaces, making raw spectral coordinates solver-dependent and therefore unreliable as node features across different meshes.

GIST resolves both problems simultaneously. It avoids eigendecomposition by computing spectral embeddings via random projection \citep{chen2019fastrp}, and it derives attention weights exclusively from gauge-invariant quantities.
Given a spectral filter $f$ and a random matrix $R \in \mathbb{R}^{N \times r}$ with $\mathbb{E}[RR^\top] = I$, each node $i$ is embedded as the $i$-th row of $\tilde{\Phi} = f(P) \cdot R \in \mathbb{R}^{N \times r}$, where $P = D^{-1}A$ is the random-walk matrix of $G$ and $r$ is the embedding dimension.
Attention weights are computed from inner products $\langle \tilde{\phi}_i, \tilde{\phi}_j \rangle$, which are unbiased estimators of the gauge-invariant kernel
\begin{equation*}
    K(i,j) = \bigl[f(P)f(P)^\top\bigr]_{ij} = \sum_{\ell} f(\lambda_\ell)^2\, (u_\ell)_i\,(u_\ell)_j,
\end{equation*}
where $\lambda_\ell$ and $u_\ell$ are the eigenvalues and eigenvectors of $L$, respectively.
Since $K(i,j)$ depends only on the operator $P$ and not on any particular eigenvector basis, it is immune to sign flips and eigenspace rotations.

As a direct consequence, GIST admits a formal bound on its discretization mismatch error.
For two discretizations $G_n$ and $G_{n'}$ of the same compact $m$-dimensional Riemannian manifold with $n \leq n'$ nodes, the attention kernel values at corresponding mesh points satisfy:
\begin{equation*}
    \bigl|\langle \tilde{\phi}_i^{n}, \tilde{\phi}_j^{n} \rangle - \langle \tilde{\phi}_i^{n'}, \tilde{\phi}_j^{n'} \rangle\bigr|
    = \mathcal{O}\!\left(n^{-\frac{1}{m+4}}\right) + \mathcal{O}\!\left(r^{-\frac{1}{2}}\right).
\end{equation*}
The first term reflects spectral convergence to the Laplace-Beltrami operator and vanishes as the mesh is refined; the second is a statistical error independently controlled by $r$, the size of the random projection space.
This quantified rate is the key theoretical guarantee absent from prior approaches: GIST does not merely claim mesh-invariance heuristically but bounds the error explicitly.
Finally, computing $\tilde{\Phi}$ via iterative power methods avoids eigendecomposition and costs $\mathcal{O}(N \cdot r)$, yielding linear scaling with mesh size that makes GIST practical on the ${\sim}4.8$M-point surface meshes in our dataset.
For a more detailed description of the method and the full mathematical formulation we refer the reader to \citet{rigotti2026gist}.

\section{Results}

\begin{figure*}[h]
    \centering
    \begin{subfigure}[b]{0.24\textwidth}
        \includegraphics[width=\linewidth]{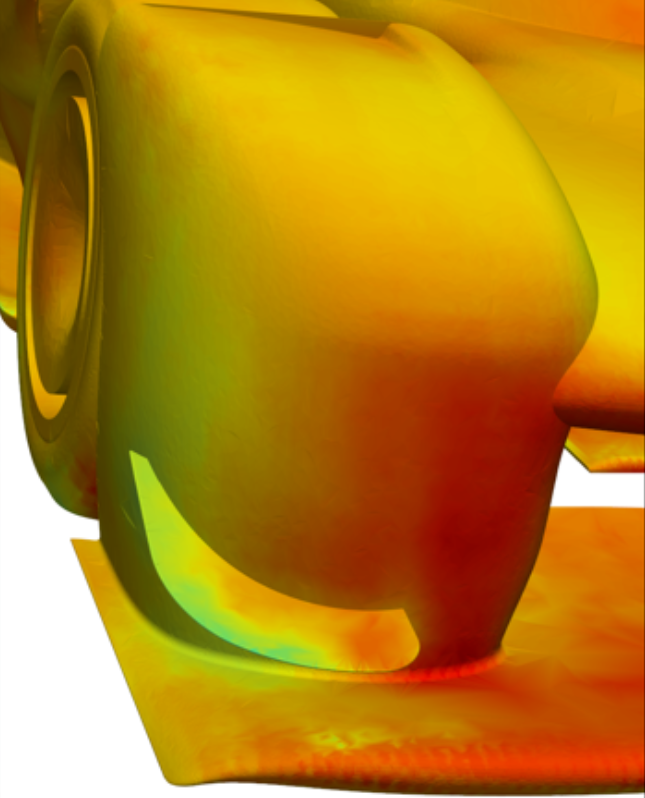}
        \includegraphics[width=\linewidth]{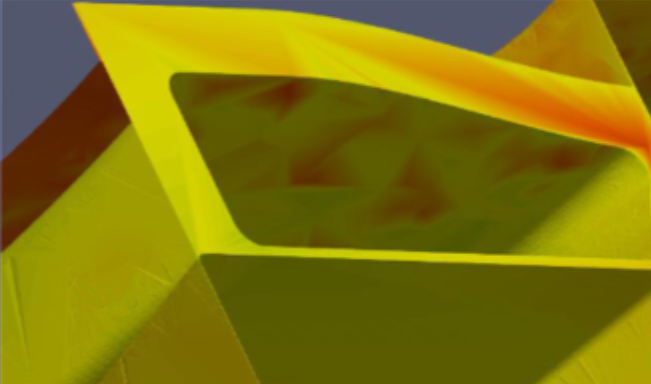}
        \caption{FNO-Based Method}
    \end{subfigure}
    \hfill
    \begin{subfigure}[b]{0.24\textwidth}
        \includegraphics[width=\linewidth]{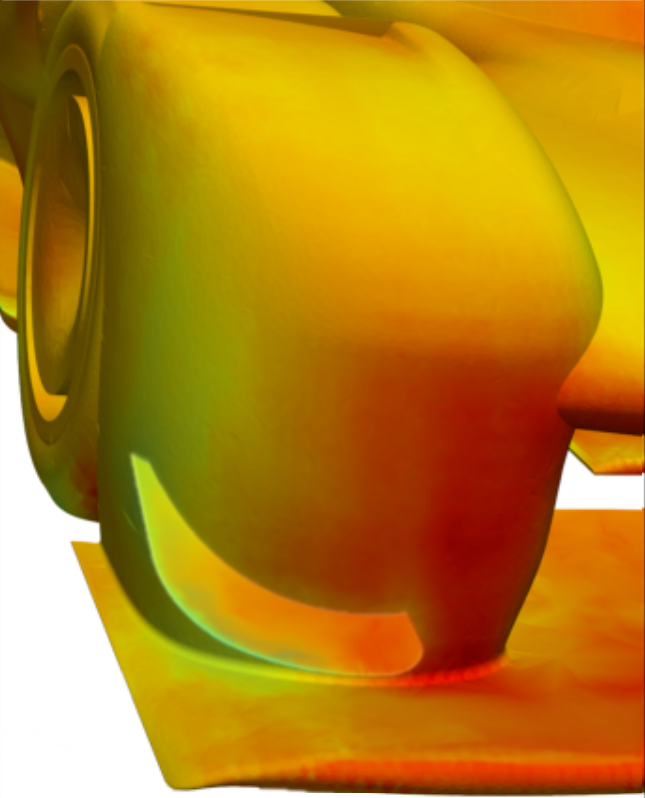}
        \includegraphics[width=\linewidth]{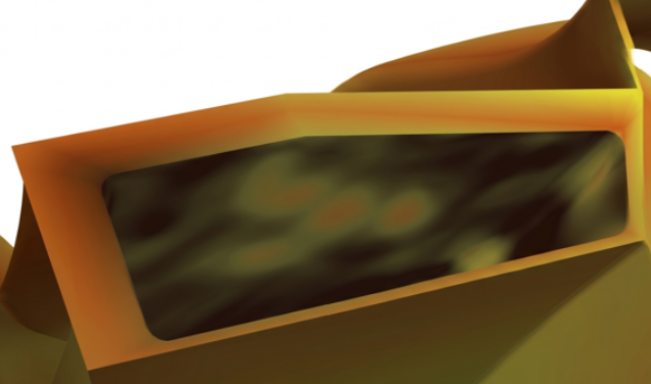}
        \caption{Transolver-Based Method}
    \end{subfigure}
    \hfill
    \begin{subfigure}[b]{0.24\textwidth}
        \includegraphics[width=\linewidth]{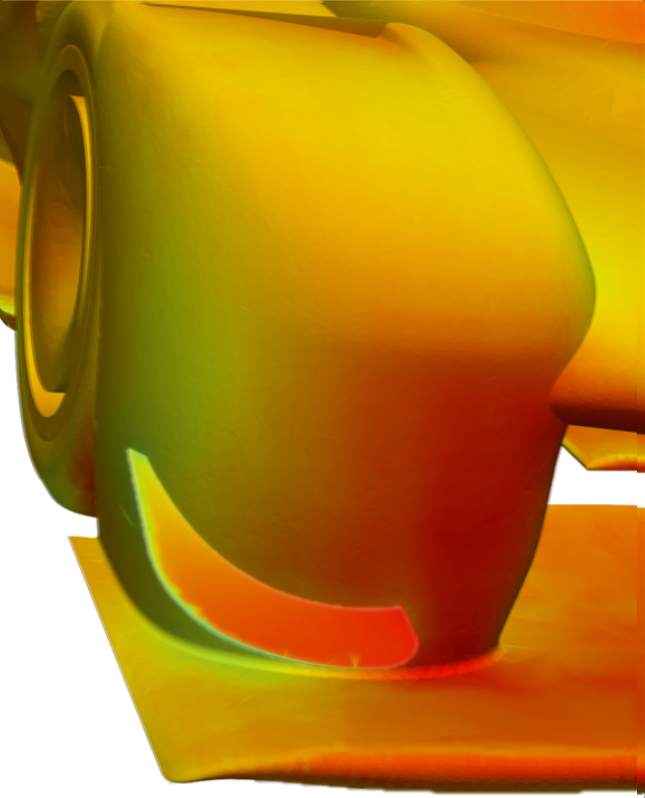}
        \includegraphics[width=\linewidth]{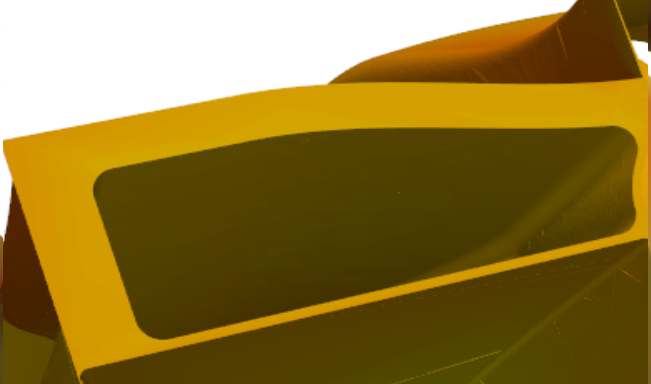}
        \caption{GIST}
    \end{subfigure}
    \hfill
    \begin{subfigure}[b]{0.24\textwidth}
        \includegraphics[width=\linewidth]{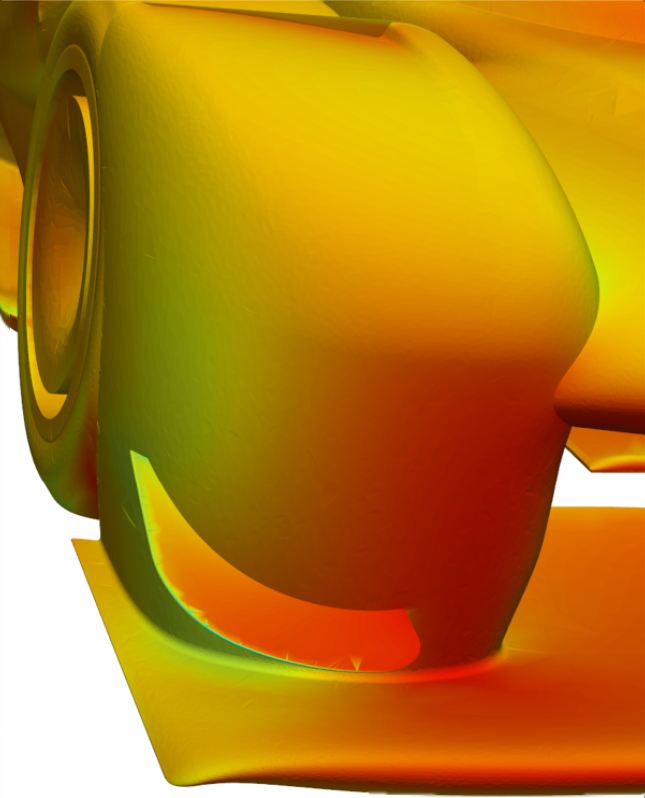}
        \includegraphics[width=\linewidth]{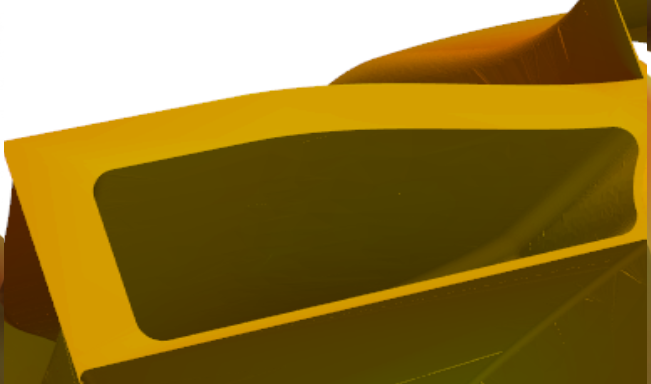}
        \caption{Ground Truth}
    \end{subfigure}
    \caption{Qualitative comparison of the pressure field predicted by different AI-surrogate models on the front area (top row) and on the rear area (bottom row) of the LMP2 race car. The pressure calculated by the GIST model is the closest to the reference CFD ground truth.}
    \label{fig:model_comparison}
\end{figure*}

The scarcity of publicly available, complex CFD datasets has hindered the community from identifying core limitations in existing neural operator approaches. A fundamental issue is the common treatment of meshes as simple point clouds, which neglects underlying mesh connectivity. Recent architectures, such as those by \cite{alkin2025ab} and \cite{wu2024transolver}, rely on the assumption that Euclidean proximity and local metadata (e.g., surface normals) provide a sufficient representation of geometry. However, for complex race-car geometries featuring thin components, two vertices may be spatially proximate yet topologically distant. A primary example is a pair of vertices located on opposite sides of a thin spoiler or flick.

Figure \ref{fig:model_comparison} illustrates this phenomenon. The Fourier Neural Operator (FNO) based model utilizes a signed distance function as input to the FNO layer and a graph neural operator (GNO) as a decoder back onto the mesh. This approach introduces a regular error pattern reflecting the underlying FNO grid (visible on the front flick) and produces unphysical predictions on the surface due to interference from invalid points in the regular grid. Similarly, while GAOT and Transolver-based methods produce smoother field predictions, they suffer from poor accuracy on thin surfaces, likely due to unphysical interactions between vertices on opposing sides. By constructing vertex embeddings from the spectral structure of the mesh graph via the gauge-invariant kernel $K(i,j)$, GIST ensures that points close in Euclidean space maintain a large embedding difference if they are topologically distant, precisely the failure mode exposed by thin race-car components such as spoilers and flicks. Consequently, GIST retains the flexibility of a point-cloud representation while rigorously treating the surface as a manifold mesh.

As a result of these considerations and the favorable scalability properties of GIST, significant performance gains are obtained over other methods.
To evaluate said performance, we assess both point-wise field accuracy and the resulting integrated aerodynamic loads when inference is run on every point in the geometry. The primary objective of the model is to predict the surface pressure $p$ and wall shear stress $\tau$ at every vertex. We quantify the error using Mean Squared Error (MSE) and the Coefficient of Determination ($R^2$):
\begin{equation*}
    \text{MSE} = \frac{1}{N} \sum_{i=1}^N (y_i-\hat{y}_i)^2, \quad R^2 = 1 - \frac{\sum_i (y_i - \hat{y}_i)^2}{\sum_i (y_i - \bar{y})^2}
\end{equation*}
where $N$ is the number of vertices of the surface mesh, $y_i$ and $\hat{y}_i$ represent the ground truth (CFD) and predicted values at vertex $i$, respectively, and $\bar{y}$ is the mean of the ground truth values.

Furthermore, the surrogate model must accurately reproduce the integrated loads used in race car development. We compute six aerodynamic coefficients by integrating force and torque contributions over the surface mesh.
Without going into details, assuming the surface mesh being discretized in surface mesh elements, the force acting on element $k$ is approximated as:
\begin{equation*}
    F_k = (\bar{p}_k n_k + \bar{\tau}_{k,\text{wall}}) A_k
\end{equation*}
where $\bar{p}_k$ and $\bar{\tau}_{k,\text{wall}}$ are representative static pressure and wall shear stress vector, respectively, computed from the point-wise predicted values on the vertices of the element $k$, $n_k$ is the outward-facing normal and $A_k$ is the area.
To resolve these forces into the aerodynamic frame, we utilize the wind-aligned coordinate system. 
The aerodynamic coefficients for drag ($C_xS$), side force ($C_yS$), and downforce ($C_zS$) are calculated by normalizing the integrated contributions:
\begin{equation*}
    C_{\{x,y,z\}}S = \frac{\sum_k (F_k \cdot e_{\{x,y,z\},\text{wind}})}{0.5 \rho_\infty V_\infty^2},
\end{equation*}
where $\rho_\infty = 1.225 \text{ kg/m}^3$ and $V_\infty = 50 \text{ m/s}$. Torque coefficients ($C_{mx}S, C_{my}S, C_{mz}S$) are similarly integrated as:
\begin{equation*}
    C_{m\{x,y,z\}} S = \frac{\sum_k (M_k \cdot e_{\{x,y,z\},\text{wind}})}{0.5 \rho_\infty V_\infty^2 L_{ref}}
\end{equation*}
where $L_{ref}$ is the characteristic length and $M_k = (\bar{r}_k - r_0) \times F_k$ is the torque of element $k$ relative to the origin $r_0$. 
For this paper, we present for simplicity the drag and downforce aerodynamic coefficients.

\subsection{Model results}

\begin{figure*}[!ht]
\centering
\begin{tabular}{p{\textwidth}}
\multicolumn{1}{c}{
} \\
\includegraphics[width=\textwidth]{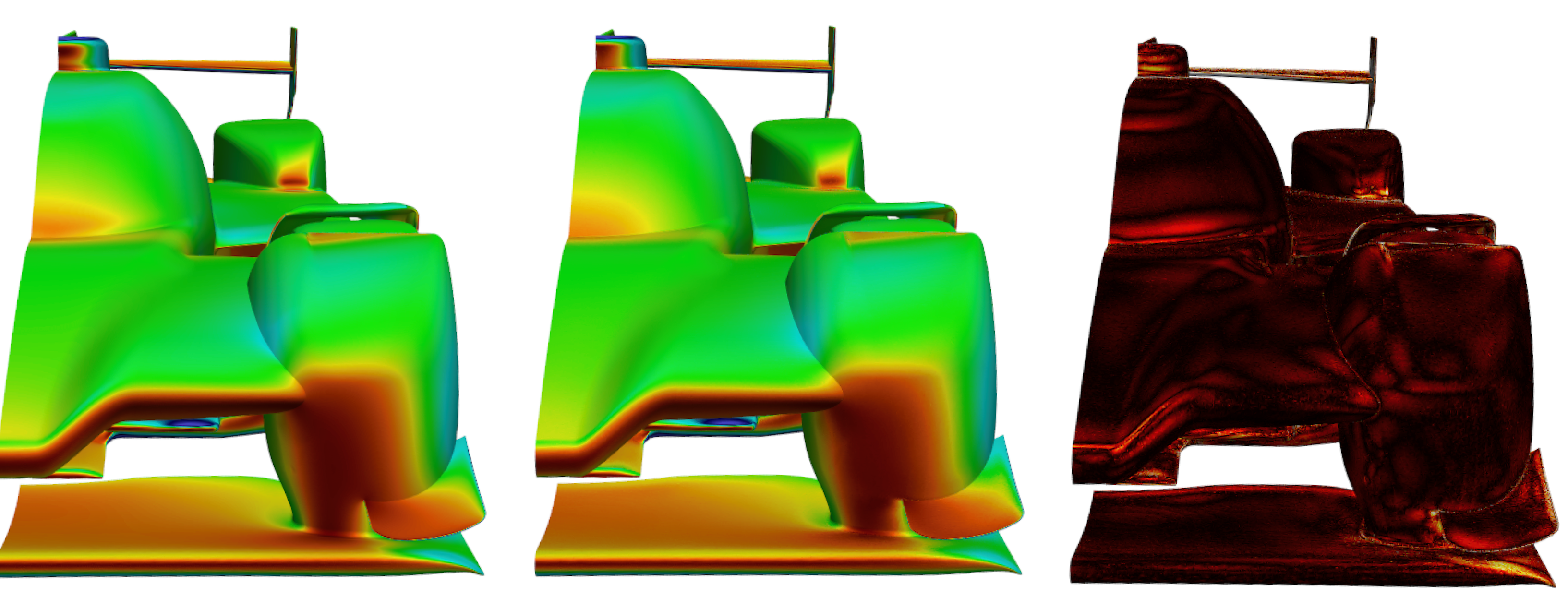}\\
\multicolumn{1}{c}{
} \\
\includegraphics[width=\textwidth]{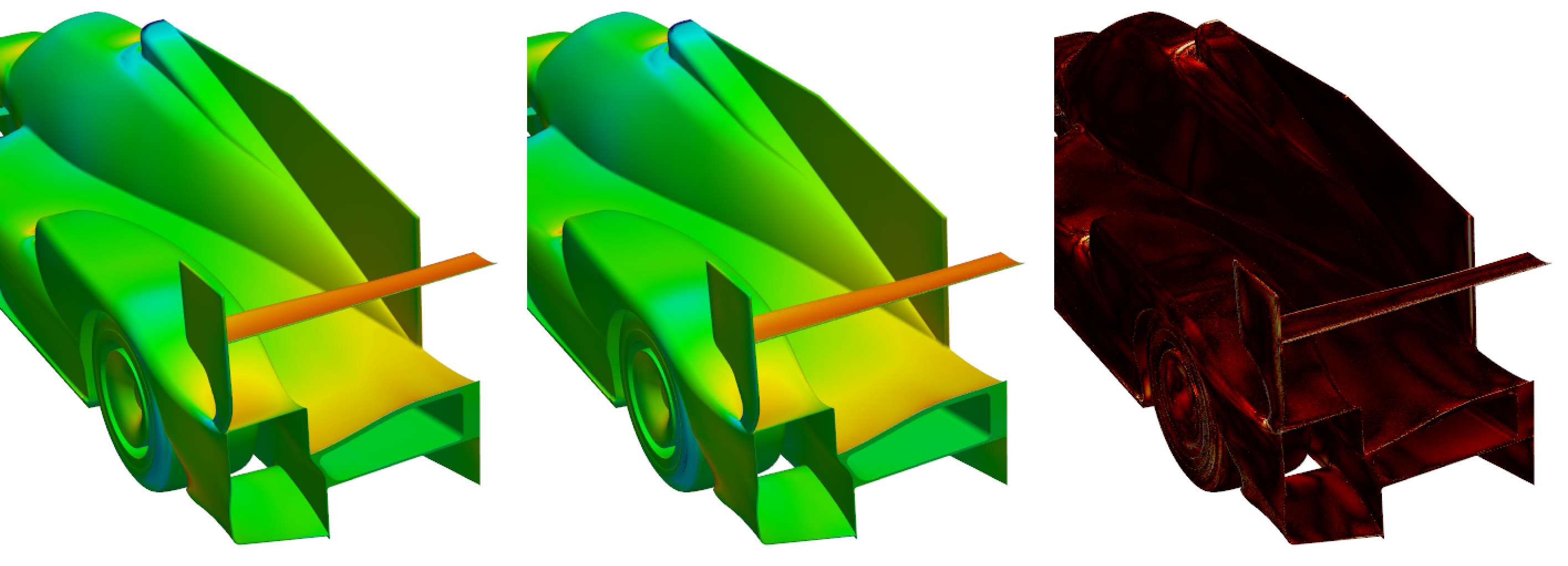}
\end{tabular}
\caption{Pressure field distribution on the complete LMP2 race car. Top row: front view. Bottom row: back view. Left: CFD ground truth. Center: GIST prediction. Right: delta.}
\label{fig:full_page_visualizations}
\end{figure*}

\begin{table}[!tb]
    \footnotesize
    \centering
    \caption{Performance comparison on the Dallara LMP2 dataset. Metrics include MSE (units of $10^{-2}$) and $R^2$ for surface pressure, and Mean Percentage Error (\%) for integrated drag ($C_d$) and downforce ($C_l$) coefficients. Higher is better for $R^2$; lower is better for all other metrics.}
    \label{tab:dallara_results_percentage}
    \begin{tabular*}{\columnwidth}{@{\extracolsep{\fill}} l cc cc @{}}
        \toprule
        & \multicolumn{2}{c}{\textbf{Pressure} ($p$)} & \multicolumn{2}{c}{\textbf{Integral Coeffs} (\% error)} \\
        \cmidrule(r){2-3} \cmidrule(l){4-5}
        \textbf{Model} & \textbf{MSE $\downarrow$} & \textbf{$R^2$ $\uparrow$} & \textbf{$C_d$ $\downarrow$} & \textbf{$C_l$ $\downarrow$} \\
        \midrule
        FNO-based    & 6.82          & 0.968          & 4.0\%          & 3.6\%          \\
        Transolver          & 4.12          & 0.980          & 1.2\%          & 1.2\%          \\
        GAOT                & 3.55          & 0.983          & 0.9\%          & 1.0\%          \\
        \textbf{GIST (ours)} & \textbf{1.88} & \textbf{0.987} & \textbf{0.6\%} & \textbf{0.7\%} \\
        \bottomrule
    \end{tabular*}
\end{table}

The results in Table \ref{tab:dallara_results_percentage} demonstrate GIST's superiority over existing state-of-the-art methods. Baselines such as GAOT and GINO require significant downsampling of the input data via random subsampling to accommodate GPU memory constraints during training. While such downsampling may be negligible for simpler datasets like ShapeNet or DrivAerNet, it has a measurable impact on performance for complex race-car geometries characterized by sharp features and intricate perturbations. In contrast, GIST’s ability to treat the mesh as a connected graph preserves these critical features while simultaneously improving inference efficiency at test time.

From a qualitative standpoint, the model accurately predicts general pressure trends even within high-complexity regions. Consistent with many data-driven methods, GIST exhibits a tendency to smooth high-frequency features, such as steep pressure gradients or localized flow structures resulting from indirect geometric coupling.

For instance, on the upper surface of the splitter—where a vortex is generated via aerodynamic interaction—the model correctly identifies the vortical structure but under-predicts the sharpest pressure gradients at the wingtip. Nevertheless, the surrogate's ability to capture the global solution profile and the presence of critical aerodynamic features provides sufficient fidelity for rapid design exploration and iterative optimization.

\begin{figure}[!htb]
    \centering
    \includegraphics[width=1\columnwidth]{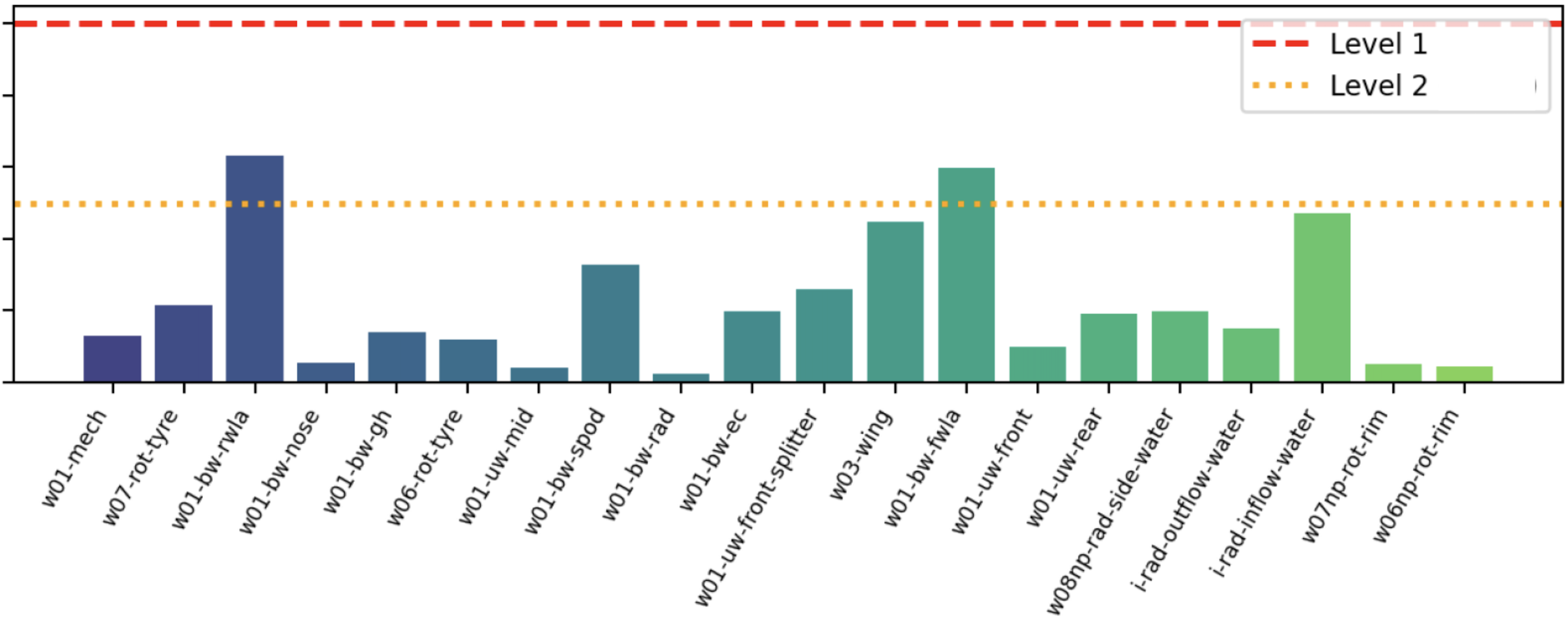}
    \caption{Absolute drag coefficient error calculated as the difference between the GIST model prediction and the CFD ground truth on each of the twenty car PIDs. The usability threshold is met for every part, while the CFD replacement threshold is met by 18/20 parts.}
    \label{fig:drag_thresholds}
\end{figure}

To evaluate the model's practical utility, Dallara established specific absolute error thresholds in advance for integrating AI into their CFD simulation pipeline. Figure \ref{fig:drag_thresholds} illustrates the average absolute error for the predicted drag coefficient across each component of the vehicle. The red line denotes the ``usability threshold'', indicating the accuracy level required for designers to reliably use the surrogate for rapid concept exploration and configuration testing. The yellow line represents a more stringent ``CFD-replacement threshold'', at which the surrogate could potentially substitute for high-fidelity simulations in specific design phases. While the specific numerical values are redacted for confidentiality, the results confirm that GIST achieves a performance level suitable for immediate productive use within Dallara’s industrial workflow. The objective is not to replace classical numerical solvers entirely, but to significantly accelerate iterative design cycles.

\subsection{Study on the optimization of rear diffuser}

\begin{figure}[!b]
    \centering
    \includegraphics[width=1.19\columnwidth]{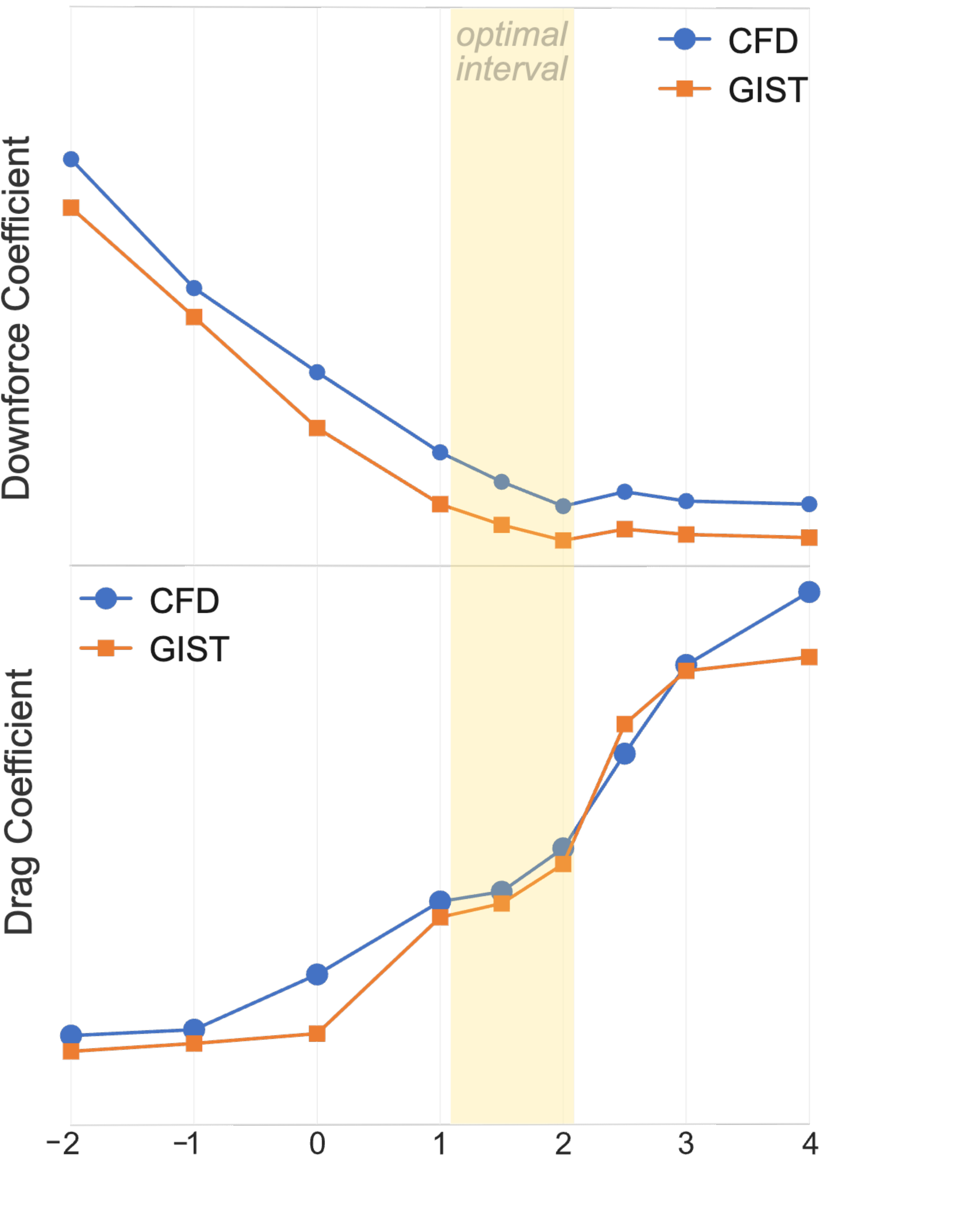}
    \caption{Downforce (top) and drag (bottom) coefficients as a function of rear diffuser angle. GIST predictions (orange) closely track CFD ground truth (blue) across the full sweep range. The shaded region highlights the optimal interval ($1$--$2^\circ$), where the downforce-to-drag ratio is maximized before downforce saturates. Note that downforce is a negative force, thus lower values mean more downforce. Values at $4^\circ$ are out of domain for the AI model: no training samples were produced near that value.}
    \label{fig:diffuser_sweep}
\end{figure}

A key use case for a trained surrogate is rapid parameter sweeps that would be highly expensive with a full CFD campaign.
To illustrate this, we study the effect of rear diffuser angle on the two primary design objectives: drag ($C_d$) and downforce ($C_l$).
Figure~\ref{fig:diffuser_sweep} shows both coefficients predicted by GIST alongside CFD ground-truth evaluations across the full range of diffuser angles, demonstrating strong agreement between the surrogate and the solver.

The two objectives exhibit competing trends.
Drag increases monotonically with angle across the entire range.
Downforce magnitude, by contrast, grows steeply for negative and small positive angles but saturates beyond approximately $1.5$--$2^\circ$: increasing the angle further delivers diminishing additional downforce.
This saturation creates a natural efficiency knee, a region where the aerodynamic gain per unit drag penalty is maximized.
A designer can formalize this as optimizing the efficiency $E=|C_l|/C_d$, or equivalently, maximizing $|C_l|$ subject to a drag budget imposed by circuit requirements or regulations.
Both formulations identify an optimal angle in the $1$--$2^\circ$ range, beyond which added downforce no longer justifies the drag cost.

Crucially, GIST produces this full sweep in seconds on a single GPU, compared to the tens of thousands of core-hours an equivalent CFD campaign would require.
This acceleration transforms diffuser angle selection from an offline batch study into an interactive design exploration, enabling engineers to respond to regulation changes or new circuit requirements in near real time.

\subsection{Next steps}
The Dallara dataset enables research trajectories impossible with current public benchmarks. Our focus is moving from static prediction to a dynamic, physics-aware design tool by leveraging the unique global and local boundary conditions inherent in our data. Unlike standard datasets that focus on fixed operating conditions, our work utilizes ``map points'' --- simulations covering varying heave, pitch, yaw, roll, and steer at multiple wind speeds. This allows for training models conditioned on diverse flow regimes, predicting aerodynamic performance during complex transient maneuvers like high-speed cornering or heavy braking.

Our most immediate and achievable objective is the integration of internal cooling aerodynamics. While academic datasets typically treat car bodies as hollow shells, each model in our dataset includes a fully functional internal radiator modeled via Darcy-flow equations. Because these porous medium labels are already present in our data, we can uniquely train the surrogate to account for ``cooling drag'' and its impact on the performance. This provides a real-time decision-making tool for the trade-off between thermal management and aerodynamic efficiency—a feature set currently unsupported by any existing AI models in the literature.

Principled uncertainty quantification is another promising direction. Equipping GIST with epistemic uncertainty estimates \citep{frick2024mc} would identify regions of the design space where predictions are less reliable, while distribution-free conformal prediction intervals \citep{deutschmann2024adaptive} could provide statistically guaranteed coverage on integrated loads, giving engineers a rigorous criterion for when to defer to the full CFD solver.

Embedding interpretability into the GIST transformer, for instance through concept-based attention mechanisms \citep{rigotti2022concept}, could further ground predictions in human-readable aerodynamic concepts and surface non-obvious cross-component couplings, deepening engineer trust and guiding targeted design exploration.

Finally, we are leveraging the differentiability of the GIST architecture to develop an Inverse Modeling framework. By propagating gradients from target performance coefficients back to input geometries, the model can automatically suggest optimal tweaks to components, from the diffuser angle to more complex and nonlinear areas. To ensure industrial reliability, we aim to bridge the ``sim-to-real'' gap using a multi-fidelity approach, fine-tuning the RANS-trained model with sparse, high-accuracy wind tunnel and track data. This calibration allows the surrogate to correct for inherent RANS limitations, such as flow separation inaccuracies, aligning AI predictions with the physical reality measured on the ground.

\section{Conclusions}
This work addresses the computational bottleneck of motorsport aerodynamic development by introducing a high-fidelity, expert-validated simplified-LMP2 dataset and the Gauge-Invariant Spectral Transformer (GIST). By explicitly encoding mesh connectivity through spectral embeddings, GIST overcomes the limitations of point-cloud-based surrogates, accurately capturing the complex flow physics of thin, highly loaded race-car components. Our results demonstrate that GIST not only achieves state-of-the-art performance on public benchmarks but also meets the rigorous accuracy thresholds required for industrial deployment at Dallara. Ultimately, this work provides a scalable, discretization-invariant foundation for the next generation of aerodynamic design tools, transitioning from passive field prediction toward dynamic, physics-aware optimization and real-time ``sim-to-real'' calibration.

\footnotesize
\bibliography{references}

@inproceedings{chen2019fastrp,
  title={Fast and Accurate Network Embeddings via Very Sparse Random Projection},
  author={Chen, Haochen and Sultan, Syed Fahad and Tian, Yingtao and Chen, Muhao and Skiena, Steven},
  booktitle={Proceedings of the 28th ACM International Conference on Information and Knowledge Management},
  year={2019}
}

@inproceedings{rigotti2022concept,
  title={Attention-based Interpretability with Concept Transformers},
  author={Rigotti, Mattia and Miksovic, Christoph and Giurgiu, Ioana and Gschwind, Thomas and Scotton, Paolo},
  booktitle={International Conference on Learning Representations},
  year={2022},
  url={https://openreview.net/forum?id=kAa9eDS0RdO}
}

@article{frick2024mc,
  title={MC Layer Normalization for calibrated uncertainty in Deep Learning},
  author={Frick, Thomas and Antognini, Diego and Giurgiu, Ioana and Grewe, Benjamin and Malossi, Cristiano and Zhu, Rong J.B. and Rigotti, Mattia},
  journal={Transactions on Machine Learning Research},
  year={2024},
  url={https://openreview.net/forum?id=bG3ICt3E0C}
}

@article{deutschmann2024adaptive,
  title={Adaptive Conformal Regression with Split-Jackknife+ Scores},
  author={Deutschmann, Nicolas and Rigotti, Mattia and Rodriguez Martinez, Maria},
  journal={Transactions on Machine Learning Research},
  year={2024},
  url={https://openreview.net/forum?id=1fbTGC3BUD}
}

@article{lu2021learning,
  title={Learning nonlinear operators via {DeepONet} based on the universal approximation theorem of operators},
  author={Lu, Lu and Jin, Pengzhan and Pang, Guofei and Zhang, Zhongqing and Karniadakis, George Em},
  journal={Nature Machine Intelligence},
  volume={3},
  pages={218--229},
  year={2021}
}

@inproceedings{pfaff2021learning,
  title={Learning mesh-based simulation with graph networks},
  author={Pfaff, Tobias and Fortunato, Meire and Sanchez-Gonzalez, Alvaro and Battaglia, Peter W.},
  booktitle={International Conference on Learning Representations},
  year={2021}
}

@book{pope2000turbulent,
  title={Turbulent Flows},
  author={Pope, Stephen B.},
  year={2000},
  publisher={Cambridge University Press}
}

@article{azizzadenesheli2024neural,
  title={Neural operators for accelerating scientific simulations and design},
  author={Azizzadenesheli, Kamyar and Kovachki, Nikola and Li, Zongyi and Liu-Schiaffini, Miguel and Kossaifi, Jean and Anandkumar, Anima},
  journal={Nature Reviews Physics},
  volume={6},
  pages={320--328},
  year={2024}
}

@article{kovachki2023neural,
  title={Neural operator: Learning maps between function spaces with applications to {PDEs}},
  author={Kovachki, Nikola and Li, Zongyi and Liu, Burigede and Azizzadenesheli, Kamyar and Bhattacharya, Kaushik and Stuart, Andrew and Anandkumar, Anima},
  journal={Journal of Machine Learning Research},
  volume={24},
  number={89},
  pages={1--97},
  year={2023}
}

@article{alkin2025ab,
  title={AB-UPT: Scaling neural CFD surrogates for high-fidelity automotive aerodynamics simulations via anchored-branched universal physics transformers},
  author={Alkin, Benedikt and Bleeker, Maurits and Kurle, Richard and Kronlachner, Tobias and Sonnleitner, Reinhard and Dorfer, Matthias and Brandstetter, Johannes},
  journal={arXiv preprint arXiv:2502.09692},
  year={2025}
}

@article{wen2025gaot,
  title={GAOT: Geometry Aware Operator Transformer},
  author={Wen, Shizheng and Kumbhat, Arsh and Lingsch, Levi and Mousavi, Sepehr and Zhao, Yizhou and Chandrashekar, Praveen and Mishra, Siddhartha},
  journal={arXiv preprint arXiv:2505.18781},
  year={2025}
}

@article{wu2024transolver,
  title={Transolver: A fast transformer solver for pdes on general geometries},
  author={Wu, Haixu and Luo, Huakun and Wang, Haowen and Wang, Jianmin and Long, Mingsheng},
  journal={arXiv preprint arXiv:2402.02366},
  year={2024}
}

@article{li2023geometry,
  title={Geometry-informed neural operator for large-scale 3d pdes},
  author={Li, Zongyi and Kovachki, Nikola and Choy, Chris and Li, Boyi and Kossaifi, Jean and Otta, Shourya and Nabian, Mohammad Amin and Stadler, Maximilian and Hundt, Christian and Azizzadenesheli, Kamyar and others},
  journal={Advances in Neural Information Processing Systems},
  volume={36},
  pages={35836--35854},
  year={2023}
}

@article{rigotti2026gist,
  title={GIST: Gauge-Invariant Spectral Transformers for Scalable Graph Neural Operators},
  author={Rigotti, Mattia and Thumiger, Nicholas and Frick, Thomas},
  journal={arXiv preprint arXiv:2603.16849},
  year={2026}
}

@article{li2020fourier,
  title={Fourier neural operator for parametric partial differential equations},
  author={Li, Zongyi and Kovachki, Nikola and Azizzadenesheli, Kamyar and Liu, Burigede and Bhattacharya, Kaushik and Stuart, Andrew and Anandkumar, Anima},
  journal={arXiv preprint arXiv:2010.08895},
  year={2020}
}

@techreport{menter1992improved,
  title={Improved two-equation k-omega turbulence models for aerodynamic flows},
  author={Menter, Florian R},
  institution={NASA Technical Memorandum},
  pages={13620},
  year={1992}
}

\end{document}